\documentclass[sigconf]{acmart}

\AtBeginDocument{%
  \providecommand\BibTeX{{%
    \normalfont B\kern-0.5em{\scshape i\kern-0.25em b}\kern-0.8em\TeX}}}

\usepackage{longtable}
\usepackage{graphicx}
\usepackage{multirow}
\usepackage{subcaption}
\usepackage{algorithm}
\usepackage{algorithmic}

\begin{document}

\title{Noise in Relation Classification Dataset TACRED: Characterization and Reduction}

\author{Akshay Parekh, Ashish Anand and Amit Awekar}
\email{akshayparekh, anand.ashish, awekar @iitg.ac.in}
\affiliation{%
  \institution{Indian Institute of Technology Guwahati}
  \city{Guwahati}
  \state{Assam}
  \country{India}
}


\begin{abstract}
TACRED, a large-scale crowd-sourced supervised dataset for Relation Classification, has already been shown noisy by earlier studies. A significant performance improvement was observed when instances with label noise were manually reannotated with the help of linguistic experts or crowd-sourcing with improved guidelines. However, the nature of noise in the dataset is largely unexplored. The overarching objective of this paper is two-fold. First, to explore model-based approaches to characterize the primary cause of the noise. Second, to identify the potentially noisy instances. Towards the first objective, we analyze predictions and performance of state-of-the-art (SOTA) models to identify the root cause of noise in the dataset. Our analysis of TACRED shows that the majority of the noise in the dataset originates from the instances labeled as no-relation which are negative examples. For the second objective, we explore two nearest-neighbor-based strategies to automatically identify potentially noisy examples for elimination and reannotation. Our first strategy, referred to as \textit{Intrinsic Strategy (IS)}, is based on the assumption that positive examples are clean. Thus, we have used false-negative predictions to identify noisy negative examples. Whereas, our second approach, referred to as \textit{Extrinsic Strategy},  is based on using a clean subset of the dataset to identify potentially noisy negative examples. Finally, we retrained the SOTA models on the eliminated and reannotated dataset. Our empirical results based on two SOTA models trained on TACRED-E following the \textit{IS} show on an average 4\% F1-score improvement, whereas reannotation (TACRED-R) does not improve the original results. However following \textit{ES}, SOTA models show the average F1-score improvement of 3.8\% and 4.4\% when trained on respective eliminated (TACRED-EN) and reannotated (TACRED-RN) datasets respectively. We further extended the \textit{ES} for cleaning positive examples as well, which resulted in an average performance improvement of 5.8\% and 5.6\% for the eliminated (TACRED-ENP) and reannotated (TACRED-RNP) datasets respectively.
\end{abstract}




\maketitle

\section{Introduction}
\label{sec:intro}
\begin{figure}[t]
    \centering
    \includegraphics[width=0.5\textwidth]{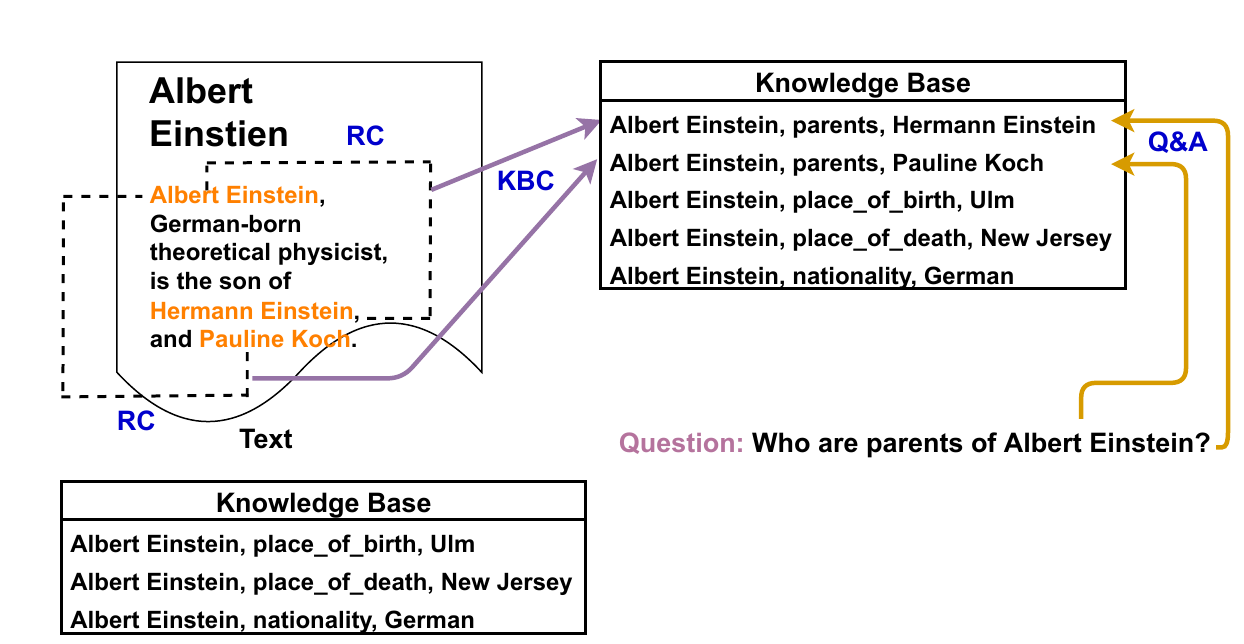}
    \caption{Example of RC and its potential applications. The relation triples (\textit{Albert Einstien}, $per:parent$, \textit{Hermann Einstien}) and (\textit{Albert Einstien}, $per:parent$, \textit{Pauline Koch}) from the sentence in the Figure, is first identified using RC and then it is populated in the Knowledge Base (KBC: Knowledge Base Completion). Populated facts are further used by question-answering (Q\&A) system.}
    \label{fig:example}
\end{figure}

Relation classification (RC), a task of identifying relations between a given pair of entities in a sentence is fundamental to Information Extraction systems. The identified structured triple (\textit{subject\_entity}, \textit{relation}, \textit{object\_entity}) from unstructured text can vastly help in knowledge base completion \cite{lin2015learning, trisedya2019neural}. This organized relational knowledge can further be used for other downstream tasks like question answering \cite{IEQA, kbqa}, and common sense reasoning \cite{kagnet}. Figure~\ref{fig:example} illustrates the RC task and one of its potential applications in KB completion and question-answering tasks.
{}
    
The TAC Relation Extraction Dataset (TACRED) \cite{tacred} is one of the most widely used benchmark datasets for large-scale RC tasks. It contains more than 100 thousand sentences labeled with one of the 41 relation classes or \textit{no\_relation}. Recent studies \cite{tacrev, retacred} have shown that the crowd-sourced TACRED contains significant annotation errors. Both these studies primarily rely on manual intervention in reannotating the partial or the complete TACRED. Although \citet{tacrev} did use multiple models to identify $5000$ most challenging or error-causing instances from the \textit{test} and \textit{dev} sets of TACRED, a subset of the identified instances was manually reannotated. On the other hand, \citet{retacred} comprehensively analysed the TACRED instances manually and modified the annotation guideline for crowd-sourced reannotation. Both studies have shown significant performance improvement on the respective revised test sets. Even though the impact of these studies is significant, primary dependence on manual intervention is a limiting factor due to the associated high cost of time and money. This motivated us to ask a few questions: \textit{(i)} Can we identify the noisy examples with a relatively small cost? \textit{(ii)} Can we perform automated reannotation with minimal human intervention?

Based on the above discussions and with the motivation to answer the above two questions, this work is aimed at
\textit{(i)} systematically studying and identifying noise present in the dataset using RC models only, \textit{(ii)} exploring model-based strategies to identify noisy instances for elimination and reannotation, and \textit{(iii)} performing comparative analyses on these strategies.

Towards the first objective, we deploy standard exploratory data analysis and machine learning tools to identify noise present in TACRED. For example, confusion matrix analysis on the test set generates a hypothesis on characteristics of misclassified instances. Another set of analyses includes downsampling, binary classification between \textit{relation} and \textit{no\_relation}, etc to differentiate between the impact of the imbalanced and noisy nature of data. All these analyses ensure an automated approach for identifying the cause of noise in the dataset. Moreover, such analyses can be done for any classification task thus making our approach applicable for any task or domain.

\begin{figure*}[t]
        \centering
        \includegraphics[width=0.8\textwidth]{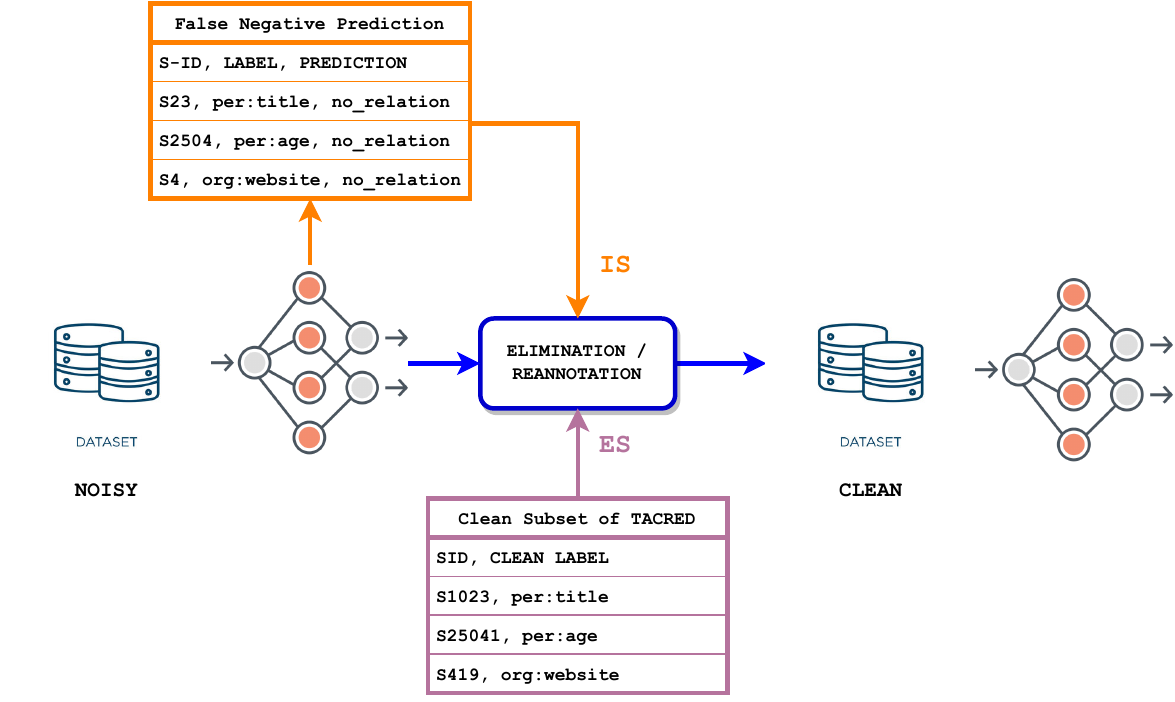}
        \caption{Pipeline for identifying noisy examples. The two strategies differ in the seed set used for identifying noisy examples. Orange (highlighted) represents using model's false negative prediction. Purple (highlighted) represents using a clean subset of TACRED. ES: Extrinsic Strategy; IS: Intrinsic Strategy}
        \label{fig:strategies}
    \end{figure*}

Once we characterize noise present in the data and understand their impact on the model performance, the next obvious questions are \textit{what to do to improve the dataset quality and how to do that}. Elimination and reannotation of noisy instances are two possible answers to the \textit{what question}. For the \textit{how to} question, we choose a model-based approach over manual intervention primarily due to its cost-, time- and labour-effectiveness, and generalizability across datasets and tasks. We explore two different strategies using the nearest-neighbour-based approach to automatically identify potential noisy examples for elimination and reannotation. Figure~\ref{fig:strategies} illustrates the general framework adopted in this study. 

    {}

The first strategy, \textit{Intrinsic Strategy (IS)}, assumes that the instances with positive relation labels are clean and noise is present in instances labeled with \textit{no\_relation}, i.e., negative labeled instances are potentially noisy. The assumption is based on the conclusion of the above-mentioned analyses. With this assumption, the first strategy uses positive labeled instances misclassified to \textit{no\_relation} class as a seed set to identify noisy negative examples. The second strategy, \textit{Extrinsic Strategy (ES)}, uses a clean subset of TACRED instances to identify noisy samples. In the absence of a budget for manual reannotation, we assume the ReTACRED test set is clean. A subset of it is chosen as clean data to identify noisy instances in the TACRED.

Extensive experimental analysis is done using two basic baseline models PALSTM \cite{tacred} and CGCN \cite{cgcn}. These two models are chosen mainly due to the following reasons: \textit{(i)} If relatively simple, yet competitive models can show performance improvement, then complex models are more likely to show improvement, and \textit{(ii)} limited computing resources are required for these models. Empirical results show that eliminating or reannotating a small set of examples using a subset of a clean dataset can significantly improve the process of finding noisy instances. Our proposed strategy ES has shown an impressive average improvement of 3.8\% and 4.4\% when trained on eliminated (TACRED-EN) and reannotated (TACRED-RN) datasets respectively and an average improvement of 5.8\% and 5.6\% for the eliminated (TACRED-ENP) and reannotated (TACRED-RNP) datasets respectively. Furthermore, manual analysis of 100 instances from TACRED-EN and TACRED-RN shows that the error rate is around 15\% only.


\textbf{Contributions of the work can be summarized as follows:}
 \begin{enumerate}
    \item Automatic or model-based characterization of noise in the TACRED dataset.
    \item Exploring automated elimination and reannotation strategies of a noisy dataset.
    Extensive experiments and analyses indicate that it is possible to reduce the time-consuming and costly process of manual reannotation of the entire dataset.
 \end{enumerate}

The rest of the paper is organised as follows: Section~\ref{sec:rw} discusses the related works under three sub-sections: relation extraction in TACRED dataset, dataset evaluation and analysis, and dataset reannotation. We have presented our analysis of baseline models' performance on TACRED to identify the cause of the noise in the dataset in Section~\ref{sec:tacred}. In Section \ref{sec:methods}, we have detailed our proposed methods for handling noisy instances in the dataset. Sections \ref{sec:experiments} and \ref{sec:results} discuss experiment design and results, along with some of the limitations of our work, respectively. Finally, in the last section (Section~\ref{sec:conclusion}), we have presented the conclusion from this work and future research directions.

\section{Related Works}
\label{sec:rw}

\subsection{Relation Extraction}
The majority of the research work on relation extraction is focused on improving learning from either distant supervised datasets or crowd-sourced annotated datasets. Distantly supervised datasets, such as \cite{nyt}, are noisy because of their dependence on incomplete knowledge bases for relation facts. Hence, the majority of efforts either involves efficient incorporation of relation information from knowledge bases \cite{kgpool, recon, ernie} or improving learning from noisy dataset using various deep neural learning \cite{lin2016neural, peng2022distantly, midtd} models.

TACRED, a large crowd-sourced Relation Extraction dataset, was proposed by \citet{tacred} in 2017. \citet{tacred} also proposed \textit{PALSTM}, position-aware attention on top of LSTM \cite{lstm}, to effectively encode the relative distances of each word in the sentence with subject and object entities. Later, several models were motivated by Graph Convolutional Network \cite{gcn} with various approaches such as using language feature dependency trees \cite{cgcn} and attention-guided GCN \cite{aggcn}. A recent GCN-based model \cite{enriched-attention} proposed an enriched attention model by utilising knowledge about the relation arguments and the shortest dependency path. However, the performance improvement was not very significant until the introduction of pre-trained language models \cite{tre, mtb, spanbert}, knowledge-aware language models \cite{knowbert, k-adapter}, and models encoding entity-specific features \cite{ernie, luke}. 

Given TACRED's noisy and imbalanced nature, a few works have been proposed recently to address these challenges. Song et al. \cite{cakd} proposed a "classifier-adaptation knowledge distillation framework (CAKD)" for relation extraction and event detection to learn from highly imbalanced datasets. Their work is motivated by the large difference in the number of positive and negative samples in the dataset. The CAKD approach led to a significant performance improvement for all the underlying models. Park et al. \cite{curriculumrc} proposed a curriculum learning-based framework based on cross-review method \cite{cross-review}. They have used pre-trained RoBERTA \cite{improvedRE}, and have achieved state-of-the-art performance on TACRED.

\subsection{Dataset Evaluation and Analysis}
Despite the use of the leading language models pre-trained on massive corpora, the performance on relation classification is restricted to around $75\%$ F1-Score\footnote{\url{https://paperswithcode.com/sota/relation-extraction-on-tacred}} \cite{tacrev}. This motivated the research community to evaluate TACRED dataset. Alt et al.\cite{tacrev} in their work, trained 49 different RE models and ranked the test and dev examples following the number of misclassifications. They selected 5000 most misclassified examples for further evaluation from linguistic experts. They identified several types of annotation errors and corrected 50\% of those examples, which led to performance improvement of an average 8\% in F1-Score. Model predictions for evaluating datasets have been used earlier, for reading comprehension \cite{RCdataseteval} and sentiment analysis \cite{SAdatasetevaluation} tasks.


In this work, we use models' predictions for TACRED analysis, but in contrast to the earlier studies, \textit{(i)} it does not involve any human expert, \textit{(ii)} it complements model predictions with several additional experiments for in-depth analysis of a dataset.

\subsection{Dataset Reannotation}
While \citet{tacrev} has highlighted the flaws in the dataset, their analysis is subjected to selected samples of test and dev sets only. So, when the supervised datasets have several annotation flaws, an obvious solution is to either detect mislabeled examples or reannotate the entire dataset. ReTACRED \cite{retacred} did a comprehensive evaluation of TACRED and employed the cost-efficient and improved annotation guidelines for crowd-sourced reannotation of TACRED. After reannotation, 22.1\% of labels differed from the original dataset, and the F1-score on the new dataset improved on an average by 13\%."

The majority of the earlier work, using the model's prediction for detecting noisy instances, focused on eliminating misclassified instances \cite{surveymisclassification}. In contrast to the existing studies, this paper focuses on the automated reannotation of RC dataset TACRED as well. To the best of our knowledge, this is the first work addressing the automated reannotation of the RC dataset TACRED.

\section{TACRED Analysis}
\label{sec:tacred}
TACRED is a sentence-level dataset where each example contains a sentence and a pair of entities. One entity represents the \textit{subject entity} and the other represents the \textit{object entity}. Each example is labeled with one of the 41 relation labels (positive classes) or \texttt{no\_relation} (negative class). Table~\ref{tab:data-split} gives the summary statistics of TACRED across the three splits of training, dev, and test sets. We can observe that the negative class accounts for almost 80\% of the entire dataset.
    
    \begin{table}[]
        \centering
        \resizebox{0.5\textwidth}{!}{%
        \begin{tabular}{cccc}
        \hline
        \textbf{Data Split} & \textbf{\# Positive} & \textbf{\#Negative} & \textbf{\# Total} \\
        \hline
        Train & 13012 & 55112 & 68124 \\ 
        Dev & 5436 & 17195 & 22631 \\ 
        Test & 3325 & 12184 & 15509 \\
        \hline
        \end{tabular}%
        }
        \caption{Number of instances across different splits of TACRED.}
        \label{tab:data-split}
    \end{table}
 
Since its introduction 4 years ago, numerous deep learning models based on LSTM \cite{tacred}, graph convolutional neural network \cite{cgcn, aggcn}, and language models \cite{spanbert, knowbert, luke} have been proposed and benchmarked on it. Despite the models' increasing complexity, performance on TACRED has been restricted to around ~75\% F1 score. In recent years, some of the works \cite{tacrev} and \cite{retacred} have attempted to analyse and re-annotate TACRED manually and have achieved significant improvements on the respective updated datasets.
    
We first analyze the performance of a few baseline models including PALSTM \cite{tacred} and CGCN \cite{cgcn}, and observe that the model's accuracy in predicting negatively labeled data is significantly high in comparison to positively labeled data (Table \ref{tab:accuracy-diff}), a common phenomenon on TACRED for other models \cite{spanbert, luke} as well. This makes us question that, "Where do models go wrong in accurately predicting the positive examples?"

    \begin{table*}[]
        \centering
        \resizebox{0.8\textwidth}{!}{%
        \begin{tabular}{|c|c|c|c|c|c|c|}
        \hline
        \textbf{Model} & \textbf{Precision} & \textbf{Recall} & \textbf{F1-score} & \textbf{Pos Acc} & \textbf{Neg Acc} & \textbf{Neg-Pos Accuracy} \\ \hline
        PALSTM & 67.8 & 64.6 & 66.2 & 64.6 & 92.6 & 28 \\ \hline
        LSTM & 65.4 & 61.6 & 63.4 & 61.6 & 92.6 & 31 \\ \hline
        BiLSTM & 65.2 & 60.4 & 62.7 & 60.4 & 92.8 & 32.4 \\ \hline
        CGCN & 71.2 & 62.6 & 66.7 & 62.6 & 94.2 & 31.6 \\ \hline
        GCN & 69.2 & 60.3 & 64.5 & 60.3 & 93.8 & 33.5 \\ \hline
        \end{tabular}%
        }
        \caption{Baseline Model performance on TACRED. Pos Accuracy and Neg Accuracy are the model's accuracy in predicting positive and negative examples respectively. }
        \label{tab:accuracy-diff}
    \end{table*}

\begin{figure*}[t]
        \begin{subfigure}[b]{0.5\linewidth}
          \centering
          \includegraphics[width=\linewidth]{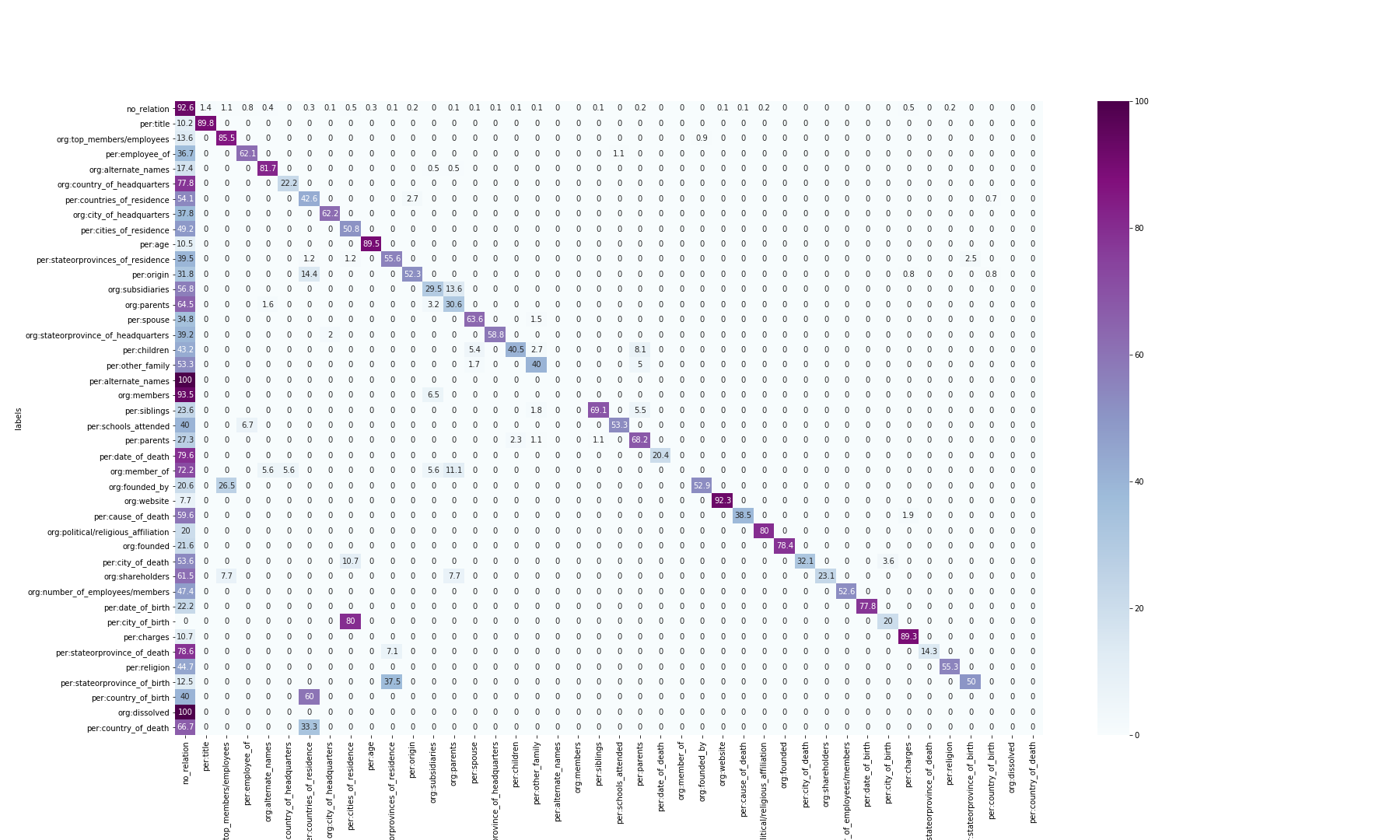}
          \label{fig:cm-parnn}
          \caption{PALSTM}
        \end{subfigure}%
        \begin{subfigure}[b]{0.5\linewidth}
          \centering
          \includegraphics[width=\linewidth]{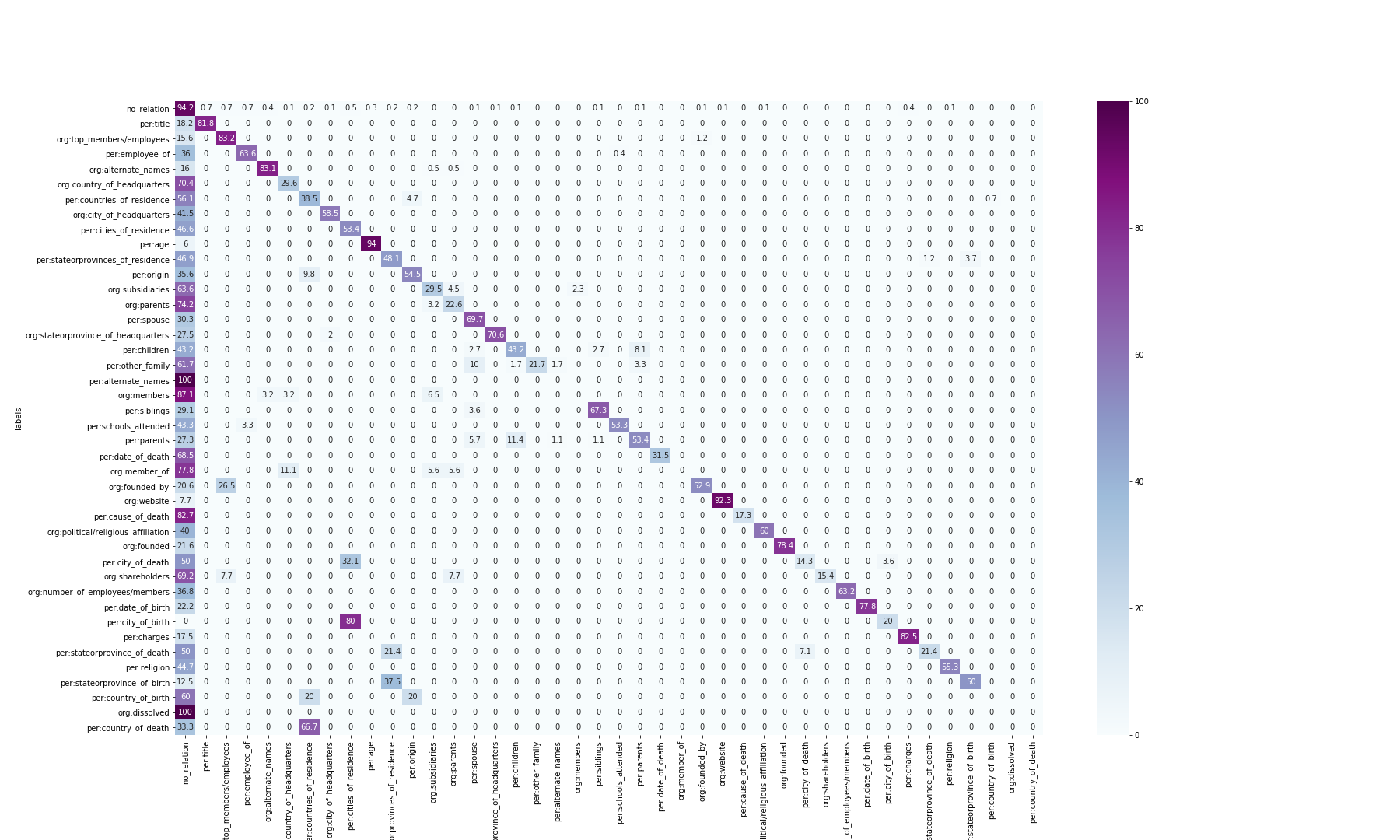}
          \label{fig:cm-cgcn}
          \caption{CGCN}
        \end{subfigure}
        \caption{Confusion Matrices for PALSTM and CGCN models; the number indicates the percentage. Rows represent the ground-truth labels and columns represent predicted labels.}
        \label{fig:cm_baseline}
    \end{figure*}

    To better understand the model's predictions in an attempt to answer the above question, we examine the prediction results using a confusion matrix. \textit{Confusion matrix} is a 2-d grid, where each row represents the predicted labels' distribution for examples belonging to a relation class, for example, in Figure~\ref{fig:cm_baseline}, row associating relation \texttt{per:title} shows that 18.2\% of examples with ground truth \texttt{per:title} in the test set is predicted as \texttt{no\_relation} by the CGCN model and remaining 81.8\% are predicted correctly. In both the confusion matrices in Figure~\ref{fig:cm_baseline}, we observe that for the considerable number of relations (20 for PALSTM and 21 for CGCN), \texttt{no\_relation} is predicted for more than 40\% of its example (entries in the first column above 40\%), which indicates either of the following:
    \begin{enumerate}
        \item The chosen models are weak and could not learn to differentiate between different classes
        \item Model's incompetence to learn due to high imbalance in the dataset.
        \item There is a significant overlap between positively labeled (one of the 41 relation classes) examples with negative examples. In other words, instances are correctly labeled but the context is very similar.
        \item There are substantial relation examples incorrectly annotated as \texttt{no\_relation}.
    \end{enumerate}

We perform further analyses to establish which among the above hypotheses are likely to be correct or have a significant impact on performance.
    
\subsection{Positive Relation Classification Analysis}
\label{subsec:prca}
\begin{figure*}[t]
        \begin{subfigure}[b]{0.5\linewidth}
          \centering
          \includegraphics[width=\linewidth]{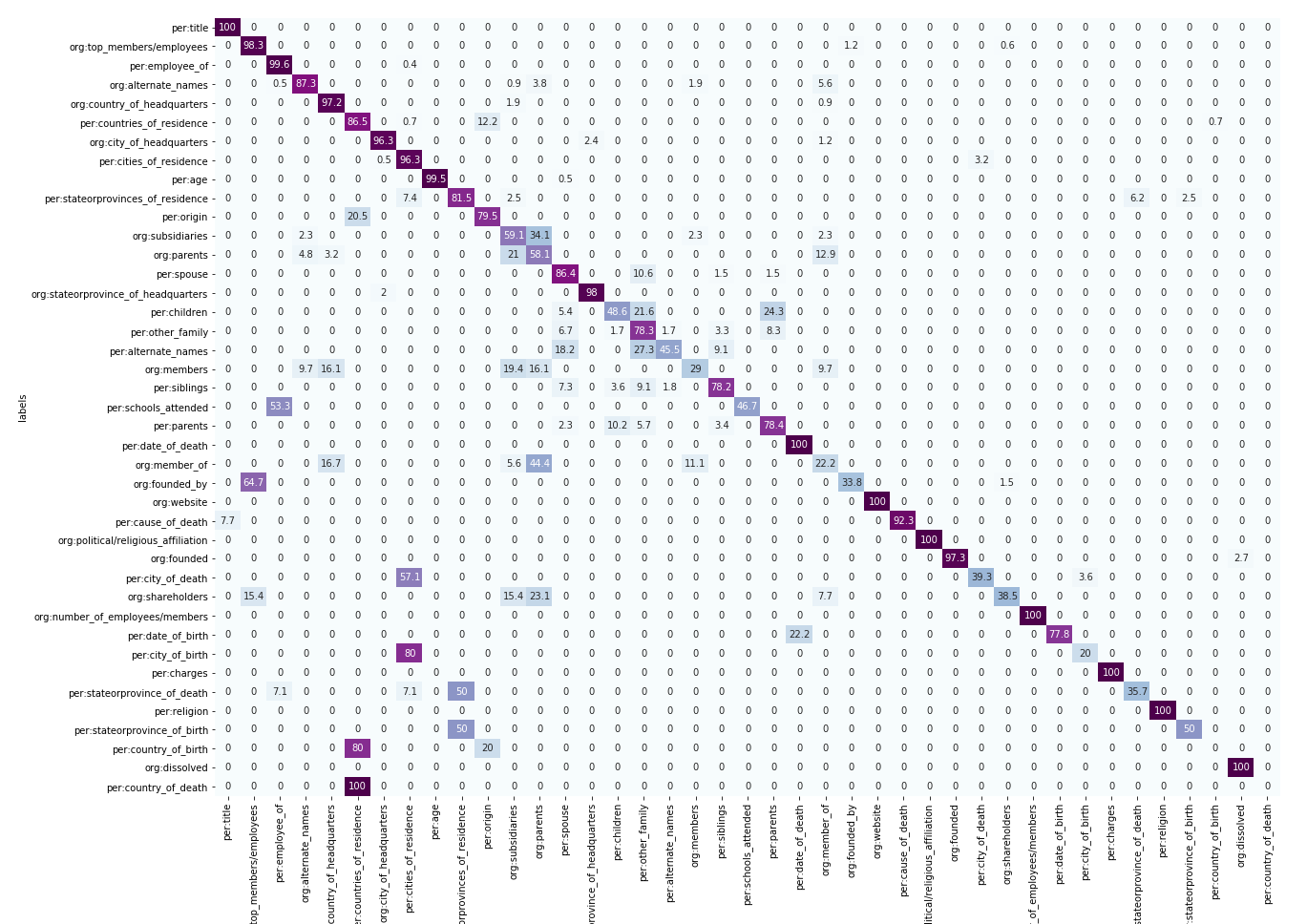}
          \label{fig:cm-parnn-allpos}
          \caption{PALSTM}
        \end{subfigure}%
        \begin{subfigure}[b]{0.5\linewidth}
          \centering
          \includegraphics[width=\linewidth]{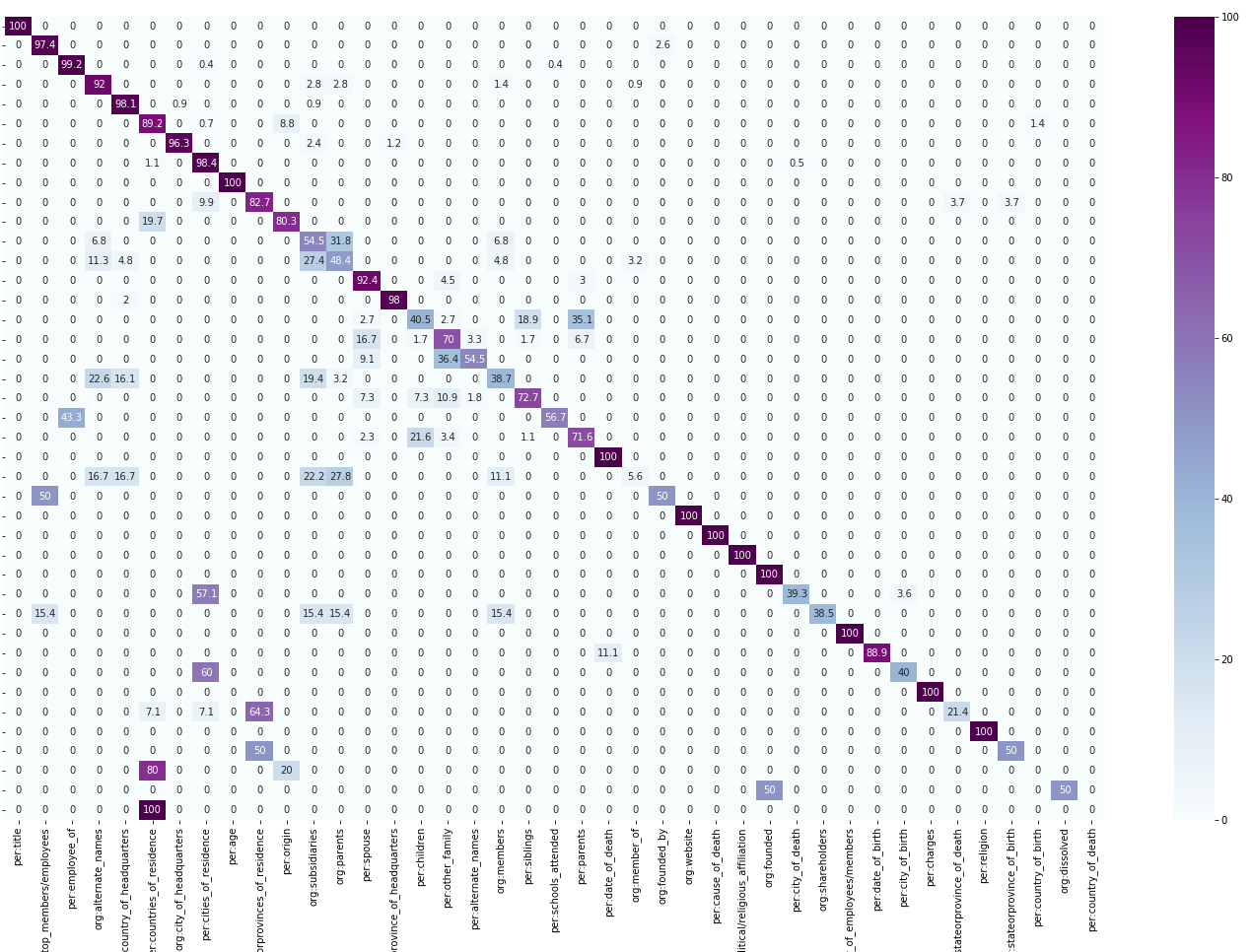}
          \label{fig:cm-cgcn-allpos}
          \caption{CGCN}
        \end{subfigure}
        \caption{Confusion Matrices for PALSTM and CGCN models excluding no\_relation; the number indicates the percentage. Rows represent the ground-truth labels and columns represent predicted labels.}
        \label{fig:cm_allpos}
    \end{figure*}

We trained the baseline models excluding \texttt{no\_relation} examples and generated a confusion matrix to comprehend the models' relation learning potential. We observe that the majority of diagonal elements (correct prediction; 29 and 31 relations for PALSTM and CGCN respectively) in Figure~\ref{fig:cm_allpos} are equal to or above 50\%. The overall performance results are shown in Table \ref{tab:all_positive}. Contrasting these results with the results (Table \ref{tab:accuracy-diff} and Figure \ref{fig:cm_baseline}) of models' performance using the entire dataset, indicates that the models are capable of learning relations despite fewer training examples and hence confirm the models' competence. There is confusion among a few (positive) relation classes, like \texttt{per:city\_of\_birth} and \texttt{per:cities\_of\_residences}; \texttt{org:founded\_by} and \texttt{org:top\_members/employee}, but they are expected as one of the relations is a subset of the other relation.
    
Based on the above discussion, we can claim that even simpler models such as PALSTM and CGCN are capable of learning relation boundaries if the dataset is relatively clean and has a balanced set of examples for all the classes. We further examine the reason behind the low performance in the following subsections with another set of analyses.

    \begin{table}[]
        \centering
        \begin{tabular}{|c|c|}
        \hline
        \textbf{Model} & \textbf{F1-Score} \\ \hline
        PALSTM          & 88.2              \\ \hline
        LSTM           & 87.2              \\ \hline
        BiLSTM         & 86.6              \\ \hline
        CGCN           & 88.7              \\ \hline
        GCN            & 87.2              \\ \hline
        \end{tabular}
        \caption{Performance of baseline models on TACRED excluding \texttt{no\_relation} examples.  Each number is reported as a percentage.}
        \label{tab:all_positive}
    \end{table}
    
\subsection{Downsampling}
\label{subsec:downsampling}
\begin{figure}[t]
        \centering
        \includegraphics[width=0.5\textwidth]{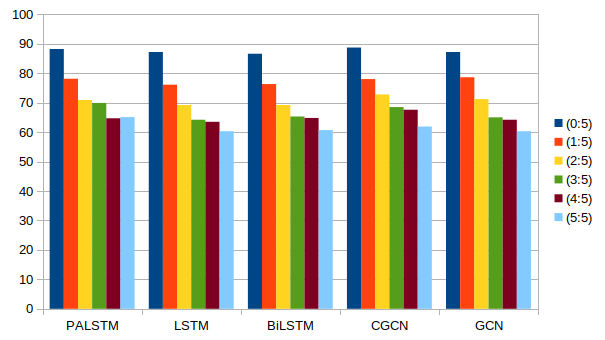}
        \caption{Effect of downsampling \texttt{no\_relation} from dataset while training on positive accuracy (Recall). 1:5 implies, that 1 out of 5, i.e. only 20\% of negative examples are selected.}
        \label{fig:downsampling}
    \end{figure}

In the previous subsection, we have seen that models are capable of learning relation boundaries among positively labeled instances in the absence of negative examples. This rules out the models' incompetence to learn efficiently. To further assess models' capabilities to learn precisely from the imbalanced data, we explore downsampling the number of negative examples. We experiment by randomly removing a portion of the negative samples from the training and development set and not modifying the evaluation dataset. If the sampling ratio is 3:5 that implies that out of every five \texttt{no\_relation} examples parsed, three are taken into the training/development set. As we move from sampling all the negative examples (ratio 5:5) to none of the examples (0:5), the accuracy of predicting positive examples (Recall) increases dramatically (Figure~\ref{fig:downsampling}).
    
On observing the trend from 0:5 to 5:5 across baselines, the effect of imbalance in data is quite apparent. However, if we consider a jump of 20\% in negative examples (i.e. 1:5 ratio), \textit{(i)} the number of positive and negative examples are almost proportionate (13012 positive examples and 11022 negative examples) and \textit{(ii)} the recall falls close to 10\% across all the baseline models. This sharp drop in the model's potential to learn from positive examples indicates that introducing negative examples brings in extra challenges apart from a disproportionate number of samples.

    {}

\subsection{Binary Classification}
\label{subsec:bc}
To further confirm the observations from downsampling, we formulated the RC on TACRED as a binary classification problem. All the positive relations examples are grouped under a single label \texttt{relation}. Table~\ref{tab:binary} shows the performance of all baseline models for binary classification.
We assume that identifying whether a relation exists or not is a simpler classification problem vs identifying the exact relation class from 41 relation labels or no relation.
However, recall improves by just 5.4\% and F1-score by 5.2\% (refer to respective columns in Table~\ref{tab:accuracy-diff} and Table~\ref{tab:binary}) on average across different baseline models. This meager change in performance hints that \textit{(i)} class imbalance may not be the prominent reason, and \textit{(ii)} there are certain negative examples that either share similar context with positive examples or there are a significant number of incorrectly labeled negative examples.

    \begin{table}[]
        \centering
        \begin{tabular}{|c|c|c|c|}
        \hline
        \textbf{Model} & \textbf{Precision} & \textbf{Recall} & \textbf{F1-score} \\ \hline
        PALSTM          & 71.6               & 68.5            & 70                \\ \hline
        LSTM           & 72.5               & 65.8            & 69                \\ \hline
        BiLSTM         & 73.4               & 66.1            & 69.6              \\ \hline
        CGCN           & 73.8               & 68              & 70.8              \\ \hline
        GCN            & 71.4               & 67.1            & 69.2              \\ \hline
        \end{tabular}
        \caption{Performance of baseline models as Binary Classification (relation or no\_relation) on TACRED. Each number is reported as a percentage.}
        \label{tab:binary}
    \end{table}
\subsection{t-SNE Plots}
\label{subsec:tsne}
\begin{figure}[t]
        \centering
        \begin{subfigure}[b]{0.5\linewidth}
          \centering
          \includegraphics[width=\linewidth]{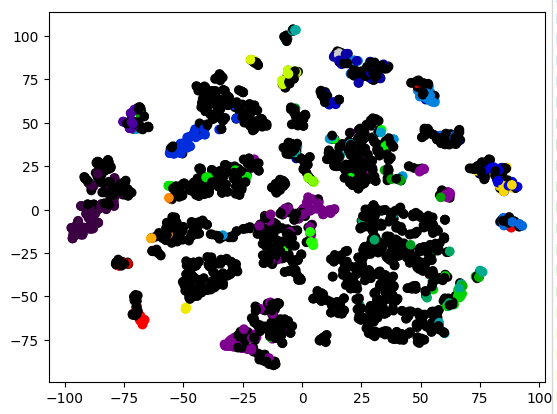}
          \caption{CGCN with no\_relation}
          \label{fig:t_cgcn}
        \end{subfigure}%
        \begin{subfigure}[b]{0.5\linewidth}
          \centering
          \includegraphics[width=\linewidth]{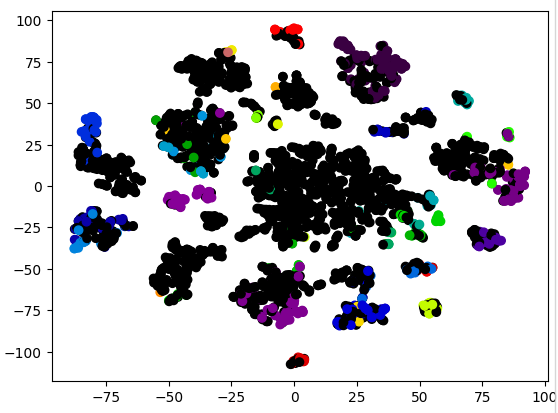}
          \caption{PALSTM with no\_relation}
          \label{fig:t_parnn}
        \end{subfigure}
        \begin{subfigure}[b]{0.5\linewidth}
          \centering
          \includegraphics[width=\linewidth]{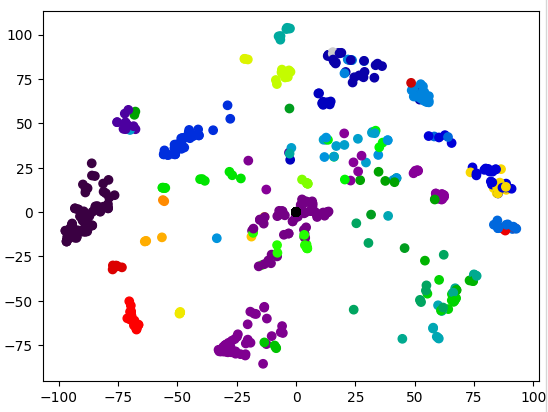}
          \caption{CGCN without no\_relation}
          \label{fig:t_cgcnf}
        \end{subfigure}%
        \begin{subfigure}[b]{0.5\linewidth}
          \centering
          \includegraphics[width=\linewidth]{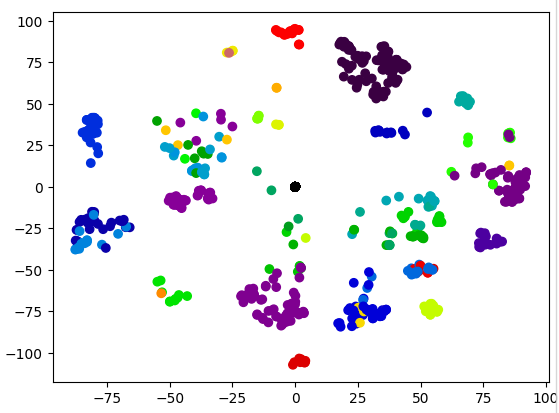}
          \caption{PALSTM without no\_relation}
          \label{fig:t_parnnf}
        \end{subfigure}
        \caption{t-SNE plots for a batch of 3000 test examples. Black dots represents instances of $no\_relation$. Different colour represents different relation label.  Refer supplementary material Figure 5 \& 6 for enlarged version.}
        \label{fig:tsne}
    \end{figure}

Above discussed analyses indicate that the presence of a disproportionate number of examples labeled with \textit{no\_relation} in the dataset may not be the prominent reason for the low performance of models. The low performance can be attributed to the nature of negative examples in the dataset. Therefore, to carefully study the characteristics of those examples, we visualize their representation using \textit{t-SNE visualization}. t-distributed stochastic neighbour embedding (t-SNE) \cite{tsne} is a non-linear dimensionality reduction algorithm for visualising high dimensional data points by mapping them to a lower dimension, preserving the information between the points. We use t-SNE plots to project embeddings learned for each relation from a random batch of 3000 test set examples to a 2-d plane. From Figures~\ref{fig:t_cgcn} and \ref{fig:t_parnn}, we observe that points (black dots) associating with \texttt{no\_relation} are all over the space which is not the case in Figures~\ref{fig:t_cgcnf} and \ref{fig:t_parnnf}, where negative examples are excluded. This shows that embeddings or representations generated by models for negative examples are quite similar to the embeddings of positive ones. This analysis indicates the likelihood of either one or more of the following cases: \textit{(i)} negative examples share context with examples from several positive classes, \textit{(ii)} negative examples do not have any specific bias towards a particular positive class and \textit{(iii)} some negative examples are potentially noisy in the sense that they are incorrectly labeled as \texttt{no\_relation}.
    
\subsection{Top-k Evaluation}
\label{subsec:topk}
In the previous section, we have analysed inconsistencies in TACRED annotation. We have shown that models are inefficient in learning from positive samples due to the noisy nature of negative examples. In this section, we further examine the same from the model's predictions by looking at the confidence score assigned to each relation for a test example. We observed that the difference between the first and second predictions is often minimal, with the second-highest predicted label frequently being the actual ground truth label. This observation motivated us to inspect the results further, and in most cases, the top two labels contain a \texttt{no\_relation} label. We hypothesized that even when the model correctly predicts a label, it gets confused by the \texttt{no\_relation} label. This observation corroborates the findings from the confusion matrix. 
    
To further empirically validate our hypothesis, we modified the evaluation criteria. If the ground truth were present in the model's $top-K$ predictions for a particular example, it would be counted as a correct prediction. In case the ground truth was not present in the $top-K$ predictions, the scorer would count it as a wrong prediction and carry out other calculations with the highest probability prediction. We re-ran experiments, taking $K = {2, 3}$. The results of the same have been outlined in Table~\ref{tab:top2} for $K = 2$ and in Table~\ref{tab:top3} $K = 3$. We have also depicted the percentage of examples where the first, second, and third predictions are correct and the percentage of examples where the ground truth is not present in any of the guesses under the columns \textit{Top1}, \textit{Top2}, \textit{Top3}, and \textit{Wrong} respectively.

    \begin{table*}[]
        \centering
        \begin{tabular}{|c|c|c|c|c|c|c|}
        \hline
        \textbf{Model} & \textbf{Precision} & \textbf{Recall} & \textbf{F1-score} & \textbf{Top 1} & \textbf{Top 2} & \textbf{Wrong} \\ \hline
        PALSTM          & 95.6               & 89              & 92.2              & 86.6           & 10.7           & 2.7            \\ \hline
        LSTM           & 93.3               & 88.3            & 90.8              & 86             & 10.9           & 3.1            \\ \hline
        BiLSTM         & 92.6               & 87.1            & 89.8              & 85.9           & 10.7           & 3.4            \\ \hline
        CGCN           & 96.1               & 89.8            & 92.9              & 87.4           & 10.2           & 2.4            \\ \hline
        GCN            & 95.7               & 88.3            & 91.8              & 86.6           & 10.6           & 2.8            \\ \hline
        \end{tabular}
        \caption{Performance of baseline modes as top k(k = 2) classifier on TACRED with percentage of top1, top2, and wrong guesses. Each number is reported as a percentage.}
        \label{tab:top2}
    \end{table*}
    
    \begin{table*}[]
        \centering
        \begin{tabular}{|c|c|c|c|c|c|c|c|}
        \hline
        \textbf{Model} & \textbf{Precision} & \textbf{Recall} & \textbf{F1-score} & \textbf{Top1} & \textbf{Top 2} & \textbf{Top3} & \textbf{Wrong} \\ \hline
        PALSTM          & 98.4               & 94.9            & 96.6              & 86.6          & 10.7           & 1.5           & 1.2            \\ \hline
        LSTM           & 96.9               & 94.2            & 95.6              & 86            & 10.9           & 1.6           & 1.4            \\ \hline
        BiLSTM         & 96.9               & 93.7            & 95.3              & 85.9          & 10.7           & 1.8           & 1.6            \\ \hline
        CGCN           & 99                 & 95.3            & 97.1              & 87.4          & 10.2           & 1.4           & 1              \\ \hline
        GCN            & 98.9               & 94.9            & 96.9              & 86.6          & 10.6           & 1.6           & 1.2            \\ \hline
        \end{tabular}
        \caption{Performance of baseline modes as top k(k = 3) classifier on TACRED with percentage of top1, top2, top3, and wrong guesses. Each number is reported as a percentage.}
        \label{tab:top3}
    \end{table*}

Based on all the above analyses (Section \ref{subsec:prca}-\ref{subsec:topk}), we conclude that
\begin{enumerate}
    \item Even the simplest model such as LSTM is capable of learning relation boundaries (Section \ref{subsec:prca}).
    \item Apart from disproportionate examples in the dataset, context overlapping and incorrect labeling of the sample as \texttt{no\_relation} is the primary concern (Section \ref{subsec:downsampling}-\ref{subsec:topk}, particularly Section \ref{subsec:tsne}).
    \item Models' performance suffers primarily due to noise originating from the negative examples in the dataset (Section \ref{subsec:bc}, \ref{subsec:topk}).
\end{enumerate}

\section{Methodology: Handling Noisy Instances}
\label{sec:methods}
Analyses in the previous section indicate that noisy instances are primarily present in the no\_relation or negative relation class. While some of the noisy instances are likely to have a similar context as one or more instances from one of the positive relation classes, the others are incorrectly labeled. This section proposes two different methodologies for identifying noisy instances. For both methodologies, we explore elimination and reannotation with appropriate labels to evaluate their impact on model performance.  We discuss the two strategies below.



\subsection{Intrinsic Strategy (IS)}
\label{subsec:false-negative}
We refer to the first strategy as \textit{intrinsic strategy (IS)}, as it utilizes only the given data and model. This strategy does not depend on the external data or the results from other models.

The intrinsic strategy assumes that \textit{substantial majority of positive samples, i.e., samples belonging to any one of the positive relation classes,  are clean}. The assumption is based on our observations that \textit{(i)} a significant number of predictions for instances from the majority of positive relation classes is \texttt{no\_relation}, and \textit{(ii)} models are capable of learning relation boundary for a majority of the positive relations if trained without no\_relation class. With this assumption, when a model $M$ makes a false-negative prediction, i.e., predicts \texttt{no\_relation} for a positive labeled sample $p_i$ with label $l_i$, it implies that there are considerable number of negative training samples ${ntr_1, ..., ntr_k} \in N_{tr}$ that either share similar context or are incorrectly labeled with \texttt{no\_relation}. 
    
Once the model is trained, we pass the target data for evaluation and collect all the instances with false-negative predictions. We refer to this collection as seed set $S$. We pass this set $S$ to the Algorithm~\ref{alg:tacred}, which uses cosine similarity for finding the $k$-nearest-neighbour from negative set $N_{tr}$. Once we have $k$ most similar negative examples ${ntr_1, ..., ntr_k} \in N_{tr}$ for each seed example $s_i \in S$, we add them to elimination list $eliminate\_sids$ if they have been not already included. Similarly, we add those examples as keys to the relabel dictionary $relabel\_sids$ with ground\_truth label $l_i$ of $s_i$ as value (relation to be reannotated with). In case, if $ntr_j$ is already present in $relabel\_sids$, we assign the label with the highest similarity score.

    \begin{algorithm}[tb]
    \caption{Algorithm for finding noisy examples for elimination and reannotation using IS}
    \label{alg:tacred}
    \textbf{Input}: Seed Set $S$\\
    \textbf{Parameter}: $k$, $N_{tr}$, $M$\\
    \textbf{Output}: $relabel\_sids$, $eliminate\_sids$ \\
        \begin{algorithmic}[1] 
        \STATE Let $t=0$. 
        \STATE Let $eliminate\_sids=list()$.
        \STATE Let $relabel\_sids=dict()$.
        \WHILE{$t < len(S)$}    
        \STATE $rel = ground\_truth(S_t)$
        \STATE $pred = M(S_t)$
        \STATE $sids = k\_similar\_negative\_examples(S_t, N_{tr}, k)$.
        \STATE Let $v=0$.
        \WHILE{$v < k$}
        \IF {$sids[v] \not\in eliminate\_sids$}
        \STATE $eliminate\_sids.append(sids[v])$.
        \STATE $relabel\_sids[sids[v]] = rel$.
        \ELSE 
        \IF {$similarity\_score(sids[v], S_t) > similarity\_score(sids[v], S_{current\_high})$}
        \STATE $relabel\_sids[sids[v]] = rel$.
        \ENDIF
        \ENDIF
        \STATE $v=v+1$.
        \ENDWHILE
        \STATE $t=t+1$.
        \ENDWHILE
        \STATE \textbf{return} $relabel\_sids, eliminate\_sids$
        \end{algorithmic}
    \end{algorithm}

\subsection{Extrinsic Strategy (ES)}

    Our first strategy of using TACRED to identify the noisy examples was based on a hypothesis that the majority of positive examples are clean. In the second strategy, we took a clean set of sentences to obtain a list of noisy examples from the TACRED. A clean subset of TACRED can be obtained by appointing some experts at a small cost. However, we took the clean set from the ReTACRED \cite{retacred} test data. As this strategy depends on the external data, we refer to this strategy as an extrinsic strategy (ES).
    
    \subsubsection{Clean Subset of TACRED}
        For reannotating TACRED, \citet{retacred} have reformulated some of the relations' definitions. Therefore, to avoid any ambiguities, we consider only those relations that can be directly mapped with the original TACRED dataset. The clean set contains 7770 negative examples and 2707 positive examples spanning across 24 relation labels. Refer to the original paper for the list of relations with updated guidelines.

    We use separate algorithms for finding noisy examples for elimination and reannotation. In both cases, we use the representation obtained from the model trained on TACRED training set. They are described next.
    
\subsubsection{Finding noisy examples for elimination}

    We use Algorithm~\ref{alg:clean_elim} for finding noisy examples for elimination. This algorithm takes the entire TACRED dataset (training, dev, and test sets) $T$ as an input. It iterates over each example $t_i \in T$ and finds $k$-nearest-neighbour from clean set $C$ using cosine similarity. Assuming the example $t_i$ has label $l_i$, the algorithm checks if none of those $k$ examples $c_1, c_2, ..., c_k \in C$ has the same label as $l_i$, it adds the example $t_i$ to the list of noisy examples for elimination $eliminate\_sids$. In other words, if label $l_i$ given in TACRED does not match with even a single $k$ nearest-neighbours from the clean set, then it is likely that $l_i$ may not be the correct label.
    
    \begin{algorithm}[tb]
    \caption{Algorithm for finding noisy examples using ES for elimination.}
    \label{alg:clean_elim}
    \textbf{Input}: TACRED Examples $T$\\
    \textbf{Parameter}: $k$, $C$\\
    \textbf{Output}: $eliminate\_sids$ \\
        \begin{algorithmic}[1] 
            \STATE Let $t=0$.
            \STATE Let $eliminate\_sids=list()$.
            \WHILE{$t < len(T)$}
                \STATE $rel = ground\_truth(T_t)$
                \STATE $sids = k\_similar\_clean\_examples(T_t, C, k)$.
                \STATE Let $v=0$.
                \STATE Let $count=0$.
                \WHILE{$v < k$}
                    \IF {$ground\_truth(sids[v]) == rel$}
                    \STATE $count = count + 1$.
                    \ENDIF
                    \STATE $v=v+1$.
                \ENDWHILE
                \IF {$count == 0$}
                \STATE $eliminate\_sids.append(T_t)$.
                \ENDIF
                \STATE $t=t+1$.
            \ENDWHILE
            \STATE \textbf{return} $eliminate\_sids$
        \end{algorithmic}
    \end{algorithm}

\subsubsection{Finding noisy examples for reannotation}

    We use Algorithm~\ref{alg:clean_reann} for finding noisy examples for reannotation. This algorithm takes TACRED dataset (training, dev, and test sets) $T$ as an input. It iterates over each example $t_i \in T$ and finds $k$-nearest neighbour from clean set $C$ using cosine similarity. Assume the example $t_i$ has label $l_i$ and its $k$-nearest-neighbours from the clean set $C$ are $c_1, c_2, ..., c_k$. The algorithm considers $t_i$ as a noisy instance if the following conditions are met:
    $ \forall j \in \{1,\ldots,k\},$ label of $c_j = l$, and $l \neq l_i$. \textbf{In other words, an instance $t_i$ is potentially noisy if all of its nearest neighbours belong to the same class but do not match with the class of $t_i$. We add $t_i$ to the dictionary of noisy examples for reannotation $relabel\_sids$, where the key is sentence id of $t_i$ and the value is $l$, the label of nearest-neighbours.}
 
    \begin{algorithm}[tb]
    \caption{Algorithm for finding noisy examples using ES for reannotation.}
    \label{alg:clean_reann}
    \textbf{Input}: TACRED Examples $T$\\
    \textbf{Parameter}: $k$, $C$\\
    \textbf{Output}: $relabel\_sids$ \\
        \begin{algorithmic}[1] 
        \STATE Let $t=0$.
        \STATE Let $relabel\_sids=list()$.
        \WHILE{$t < len(T)$}
        \STATE $rel = ground\_truth(T_t)$
        \STATE $sids = k\_similar\_clean\_examples(T_t, C, k)$.
        \STATE Let $v=0$.
        \STATE Let $rel\_lst=list()$.
        \WHILE{$v < k$}
        \IF {$ground\_truth(sids[v]) \not= rel$}
        \STATE $rel\_lst.append(ground\_truth(sids[v]))$.
        \ENDIF
        \STATE $v=v+1$.
        \ENDWHILE
        \IF {$check\_all\_relations\_are\_same(rel\_lst)$}
        \STATE $relabel\_sids[T_t] = rel\_lst[0]$.
        \ENDIF
        \STATE $t=t+1$.
        \ENDWHILE
        \STATE \textbf{return} $relabel\_sids$
        \end{algorithmic}
    \end{algorithm}

\section{Experiments}
\label{sec:experiments}

    Following the first approach discussed in subsection \ref{subsec:false-negative}, we conduct two sets of experiments, \textit{(i)} we find noisy training examples only and eliminate/reannotate them, \textit{(ii)} we find noisy examples from all three sets (train, dev, test) for elimination/reannotation. In the second set of experiments, the model is always trained using training data only.
    
    
\subsection{Baseline Models and Hyper-parameters}
    All our analyses are based on the following baseline models: PALSTM \cite{tacred}, CGCN \cite{cgcn}, LSTM, BiLSTM, and GCN. All the input word vectors are initialised using pre-trained GloVe vectors \cite{glove}. We consider 300-dimensional vectors for CGCN and GCN and 200-dimensional vectors for LSTM, BiLSTM, and PALSTM. For other tags such as POS and NER, 30-dimensional vector representations are considered.
    
    For training GCN and CGCN, we use the same set of hyper-parameters as in \cite{cgcn}. We use $200$ nodes in LSTM and feedforward hidden layers. We use 2 GCN layers and 2 feedforward layers, SGD as an optimizer, with an initial learning rate of 1.0 which is reduced by a factor of 0.9 after epoch 5. We train the model for 100 epochs. We use a word dropout probability of 0.04 and a dropout probability of 0.5 for LSTM layers. 

    For training PALSTM, LSTM, and BiLSTM, we follow \cite{tacred} for hyper-parameters. We have used 2 layered stacked LSTM layers for all the models with a hidden size of 200. We use \textit{AdaGrad} with a learning rate of 1.0 which is reduced by a factor of 0.9 after the 20th epoch. We have trained the model for 30 epochs. We use a word dropout probability of 0.04 and a dropout probability of 0.5 for LSTM layers. 
    
    
\subsection{Evaluation Models}
    Apart from the two SOTA models (PALSTM and CGCN) among baselines, we use two recent SOTA RE models based on pre-trained large-scale language models for evaluation of elimination and reannotation using ES.
    
    As recent SOTA RE models, we use SpanBERT \cite{spanbert}, which employs a bi-directional language model similar to BERT \cite{bert} pre-trained on span-level, and a recent model \cite{improvedRE} that uses pre-trained BERT along with typed entity marker to better highlight the entity information for RE. For both the methods we use \textit{base-cased} version along with the set of hyper-parameters as used in the respective works~\cite{spanbert,improvedRE}.
    
\subsection{TACRED Variant for Evaluation}

    Other than TACRED, we have also shown evaluation results on TACRev \cite{tacrev} eval data. \cite{tacrev} trained 49 different RE models to identify the most challenging examples from dev and test sets. They identified 1923 examples as challenging. 960 from that set were later reannotated by linguistic experts. This cleaner version of TACRED is referred to as TACRev by \cite{tacrev}. 


\section{Results and Discussion}
\label{sec:results}
In the following subsections, we discuss the performance of models under different experiment settings.

\subsection{Performance Evaluation}
 \textbf{Intrinsic Strategy (IS) on TACRED test set:} Results for all considered models, trained after both elimination and reannotation, on TACRED test set are presented in row 2 of each model block in Table~\ref{tab:is-elim-perf} and Table~\ref{tab:is-reann-perf} respectively. We observe that recall gets boosted significantly but at the cost of precision. This reflects that both elimination and reannotation make the model better at predicting examples from positive relation classes.
    
    
    
\noindent \textbf{IS on the cleaned TACRED test set:} In the above analysis, evaluation was done on the original TACRED evaluation set. Studies have indicated that this set is also noisy. Any evaluation of such a noisy set may not give us a true picture. To evaluate both our elimination and reannotation strategies on cleaner evaluation datasets, we performed another experiment by passing dev and test sets to the same algorithm (Algorithm~\ref{alg:tacred}). Instead of negative training samples, negative dev, and test samples were used to identify noisy examples from the two sets respectively. As discussed earlier, the model is trained using training data only. Dev and test set samples are fed to the model to obtain corresponding representations.
    
Performance of SOTA models trained after elimination and reannotation on cleaned TACRED, TACRED-E, and TACRED-R for elimination and reannotation respectively are presented in Tables~\ref{tab:is-elim-perf} and \ref{tab:is-reann-perf} respectively. The third row of each model block contains results on the original TACRED test set and the fourth row contains results on new TACRED-E (Table~\ref{tab:is-elim-perf}) or TACRED-R (Table \ref{tab:is-reann-perf}) test set.  The performance of the model went down significantly after training on reannotated TACRED. However, the model performance shows significant improvement in the F1-score after training on eliminated TACRED. Based on the result, we hypothesize that our approach of finding noisy examples using the nearest neighbours of false-negative model prediction can identify noisy examples. But, reannotated labels may not be correct and hence, the model is unable to take advantage of them.

    \begin{table*}[]
        \centering
        \resizebox{0.8\textwidth}{!}{%
        \begin{tabular}{cccccc}
        \textbf{Model} & \textbf{\# Elimination} & \textbf{Evaluation Data} & \textbf{Precision} & \textbf{Recall} & \textbf{F1-Score} \\ \hline
        \multirow{4}{*}{PALSTM} & baseline & TACRED & 67.8 & 64.6 & 66.2 \\
         & 5334, 0, 0 & TACRED & 63.6 & 65.7 & 64.7 \\
         & 5334, 3092, 0 & TACRED & 62.8 & 66.2 & 64.4 \\
         & 5334, 3092, 2038 & TACRED-E & 71.9 & 66.2 & 68.9 \\ \hline
        \multirow{4}{*}{CGCN} & baseline & TACRED & 71.2 & 62.6 & 66.7 \\
         & 5321, 0, 0 & TACRED & 66.6 & 67.4 & 67 \\
         & 5321, 3133, 0 & TACRED & 63 & 70.8 & 66.7 \\
         & 5321, 3133, 2043 & TACRED-E & 73.1 & 70.8 & 71.9
        \end{tabular}%
        }
        \caption{Performance of SOTA models after elimination of noisy negative examples identified using  Algorithm~\ref{alg:tacred} following IS on TACRED evaluation data. TACRED-E represents TACRED after the elimination of noisy examples. Except for baseline results, all the models are trained on the TACRED-E train set. \textbf{\# Elimination} represents the number of examples eliminated from (train, dev, test).}
        \label{tab:is-elim-perf}
    \end{table*}

    \begin{table*}[]
        \centering
        \resizebox{0.8\textwidth}{!}{%
        \begin{tabular}{cccccc}
        \textbf{Model} & \textbf{\# Reannotation} & \textbf{Evaluation Data} & \textbf{Precision} & \textbf{Recall} & \textbf{F1-Score} \\ \hline
        \multirow{4}{*}{PALSTM} & baseline & TACRED & 67.8 & 64.6 & 66.2 \\
         & 5334, 0, 0 & TACRED & 56.7 & 68.8 & 62.2 \\
         & 5334, 3092, 0 & TACRED & 52.8 & 73 & 61.3 \\
         & 5334, 3092, 2038 & TACRED-R & 66.7 & 57.2 & 61.6 \\ \hline
        \multirow{4}{*}{CGCN} & baseline & TACRED & 71.2 & 62.6 & 66.7 \\
         & 5321, 0, 0 & TACRED & 59.4 & 71.2 & 64.8 \\
         & 5321, 3133, 0 & TACRED & 55.4 & 73.4 & 63.1 \\
         & 5321, 3133, 2043 & TACRED-R & 68.8 & 56.5 & 62
        \end{tabular}%
        }
        \caption{Performance of SOTA models after reannotation of noisy negative examples identified using Algorithm~\ref{alg:tacred} following IS on TACRED evaluation data. TACRED-R represents TACRED after the reannotation of noisy examples. Except for baseline results, all the models are trained on the TACRED-R train set. \# Reannotation represents the number of examples reannotated from (train, dev, test).}
        \label{tab:is-reann-perf}
    \end{table*}
    
\noindent \textbf{Extrinsic Strategy(ES) on TACRED test set:} In our second strategy, we use subset of ReTACRED \cite{retacred} test set as clean dataset for identifying noisy instances using different algorithms for elimination (Algorithm~\ref{alg:clean_elim}) and reannotation (Algorithm~\ref{alg:clean_reann}). Performance of all the models trained on eliminated and reannotated negative examples of TACRED are presented in Tables~\ref{tab:es-elim-perf} and \ref{tab:es-reann-perf} respectively. The second row of each model block contains results on the original TACRED test set and the third row contains results on new TACRED-EN and TACRED-RN test sets. On both the datasets, models have shown improved performance in recall and F1-score, despite a very small fraction of updates in the TACRED.

We further evaluated this strategy by finding noisy positive examples as well, using the same algorithms. After eliminating and reannotating such examples from TACRED, the models' performance improved further (Tables~\ref{tab:es-elim-perf} and \ref{tab:es-reann-perf}). The fourth row contains results on the original TACRED test set and the fifth row contains results on the new TACRED-ENP and TACRED-RNP test sets.

    \begin{table*}[]
        \centering
        \resizebox{0.8\textwidth}{!}{%
        \begin{tabular}{c|cccccc}
        \textbf{Model} & \textbf{Train Data} & \textbf{\# Elimination} & \textbf{Evaluation Data} & \textbf{Precision} & \textbf{Recall} & \textbf{F1-Score} \\ \hline
        \multirow{5}{*}{PALSTM} & TACRED & baseline & TACRED & 67.8 & 64.6 & 66.2 \\
         & TACRED-EN & 1568, 852, 0 & TACRED & 67.1 & 66.8 & 67 \\
         & TACRED-EN & 1568, 852, 481 & TACRED-EN & 73.4 & 66.8 & 69.9 \\
         & TACRED-ENP & 2669, 1543, 0 & TACRED & 68 & 63.9 & 65.9 \\
         & TACRED-ENP & 2669, 1543, 741 & TACRED-ENP & 74.9 & 68.3 & 71.5 \\ \hline
        \multirow{5}{*}{CGCN} & TACRED & baseline & TACRED & 71.2 & 62.6 & 66.7 \\
         & TACRED-EN & 1878, 961, 0 & TACRED & 67.8 & 67 & 67.4 \\
         & TACRED-EN & 1878, 961, 466 & TACRED-EN & 74.4 & 67 & 70.5 \\
         & TACRED-ENP & 2997, 1610, 0 & TACRED & 68.2 & 66.8 & 67.5 \\
         & TACRED-ENP & 2997, 1610, 721 & TACRED-ENP & 75.2 & 70.7 & 72.9
        \end{tabular}%
        }
        \caption{Performance of SOTA  models trained after eliminating noisy examples following Algorithm~\ref{alg:clean_elim}. TACRED-E represents TACRED after the elimination of noisy examples. Except for baseline results, all the models are trained on the TACRED-E train set. \textbf{\# Elimination} represents the number of examples eliminated from (train, dev, test).}
        \label{tab:es-elim-perf}
    \end{table*}

    \begin{table*}[]
        \centering
        \resizebox{0.8\textwidth}{!}{%
        \begin{tabular}{c|cccccc}
        \textbf{Model} & \textbf{Train Data} & \textbf{\# Reannotation} & \textbf{Evaluation Data} & \textbf{Precision} & \textbf{Recall} & \textbf{F1-Score} \\ \hline
        \multirow{5}{*}{PALSTM} & TACRED & baseline & TACRED & 67.8 & 64.6 & 66.2 \\
         & TACRED-RN & 1354, 768, 0 & TACRED & 65.8 & 66.8 & 66.3 \\
         & TACRED-RN & 1354, 768, 409 & TACRED-RN & 74.1 & 67 & 70.3 \\
         & TACRED-RNP & 2284, 1351, 0 & TACRED & 67.7 & 64.9 & 66.3 \\
         & TACRED-RNP & 2284, 1351, 634 & TACRED-RNP & 75.6 & 68.3 & 71.8 \\ \hline
        \multirow{5}{*}{CGCN} & TACRED & baseline & TACRED & 71.2 & 62.6 & 66.7 \\
         & TACRED-RN & 1642, 843, 0 & TACRED & 67.9 & 66.9 & 67.4 \\
         & TACRED-RN & 1642, 843, 389 & TACRED-RN & 76 & 67.1 & 71.3 \\
         & TACRED-RNP & 2540, 1323, 0 & TACRED & 69.2 & 65.2 & 67.1 \\
         & TACRED-RNP & 2540, 1323, 573 & TACRED-RNP & 77.1 & 68 & 72.3
        \end{tabular}%
        }
        \caption{Performance of SOTA  models trained after reannotating noisy examples following Algorithm~\ref{alg:clean_reann}. TACRED-R represents TACRED after the reannotation of noisy examples. Except for baseline results, all the models are trained on the TACRED-R train set. \textbf{\# Reannotation} represents the number of examples reannotated from (train, dev, test).}
        \label{tab:es-reann-perf}
    \end{table*}

\noindent \textbf{ES on cleaned TACRED test set:} As discussed earlier, cleaning only the training set will not give us a true picture if the evaluation data is still noisy \cite{brodley1999identifying}. 
Thus, we have used cleaned evaluation data for reporting model performances. Here, we have also included two models based on pre-trained large-scale language models. Table~\ref{tab:perf-eval} presents the performance of the four SOTA models for TACRED and TACRev. Our proposed strategy ES for eliminating/reannotating noisy examples using the clean subset of the dataset has shown improvement in the F1-score across all the variants for all 4 models. We observe an average performance improvement of around 4\% for elimination and reannotation of only negatively labeled data on TACRED and 2.6\% and 1\% for elimination and reannotation of only negatively labeled TACRev data. While eliminating and reannotaing both positive and negative labeled noisy examples, we have an average improvement of 7.2\% and 6\% for elimination and reannotation respectively on TACRED and 4.2\% and 1.7\% for elimination and reannotation respectively on TACRev.
All the models got significant performance improvement, sometimes even more than 6\%, across all variants of the modified evaluation data following our proposed strategies. The BERT model remained the best-performing model even on modified evaluation data.
    
    \begin{table*}[]
        \centering
        \resizebox{0.8\textwidth}{!}{%
        \begin{tabular}{c|c|ccc|ccc|ccc|ccc}
         &  & \multicolumn{3}{c|}{PALSTM} & \multicolumn{3}{c|}{CGCN} & \multicolumn{3}{c|}{SpanBERT} & \multicolumn{3}{c}{BERT} \\
        Datset & Variant & Precision & Recall & F1-Score & Precision & Recall & F1-Score & Precision & Recall & F1-Score & Precision & Recall & F1-Score \\ \hline
        \multirow{5}{*}{TACRED} & Original & 67.8 & 64.6 & 66.2 & 71.2 & 62.6 & 66.7 & 66.4 & 66.1 & 66.3 & 71.1 & 71.4 & 71.2 \\ 
         & Eliminate-Neg & 73.4 & 66.8 & 69.9 (+3.7) & 74.4 & 67 & 70.5 (+3.8) & 77.2 & 65.5 & 70.9 (+4.6) & 79.6 & 71.3 & 75.2 (+4) \\
         & Eliminate-Neg\&Pos & 74.9 & 68.3 & 71.5 (+5.3) & 75.2 & 70.7 & 72.9 (+6.2) & 80.5 & 72.5 & 76.3 (+10) & 84.5 & 73 & 78.3 (+7.1) \\
         & Reannotate-Neg & 74.1 & 67 & 70.3 (+4.1) & 76 & 67.1 & 71.3 (+4.6) & 76.8 & 64.5 & 70.1 (+3.8) & 79 & 71.7 & 75.2 (+4) \\  
         & Reannotate-Neg\&Pos & 75.6 & 68.3 & 71.8 (+5.6) & 77.1 & 68 & 72.3 (+5.6) & 76.5 & 68.8 & 72.5 (+6.2) & 78.5 & 76.9 & 77.7 (+6.5) \\ \hline
        \multirow{5}{*}{TACRev} & Original & 73.7 & 74.8 & 74.3 & 77.5 & 72.6 & 75 & 71.3 & 75.6 & 73.4 & 75.9 & 81.1 & 78.4 \\ 
         & Eliminate-Neg & 75.4 & 76.9 & 76.1 (+1.8) & 75.9 & 76.9 & 76.4 (+1.4) & 79.3 & 75.7 & 77.5 (+4.1) & 81.1 & 81.7 & 81.4 (+3) \\ 
         & Eliminate-Neg\&Pos & 76.7 & 75.3 & 76 (+1.7) & 77 & 78.7 & 77.9 (+2.9) & 81.9 & 78.9 & 80.4 (+7) & 86.3 & 80.7 & 83.4 (+5) \\ 
         & Reannotate-Neg & 74.7 & 74.6 & 74.7 (+0.4) & 76.9 & 75.2 & 76 (+1) & 77.7 & 71.5 & 74.5 (+1.1) & 79.8 & 80 & 79.9 (+1.5) \\ 
         & Reannotate-Neg\&Pos & 76.8 & 73.9 & 75.4 (+1.1) & 78.4 & 74.5 & 76.4 (+1.4) & 77.2 & 73.2 & 75.2 (+1.8) & 79.2 & 82.5 & 80.8 (+2.4) \\ \hline
        \end{tabular}%
        }
        \caption{Evaluation of our proposed dataset elimination/reannotation strategy based on 4 models.}
        \label{tab:perf-eval}
    \end{table*}

A comparison of our proposed ES approaches with other TACRED variants is presented in Table~\ref{tab:perf-comp} for three baseline models. Despite being the least manual intervention approach, our best-performing method is not that far from both TACRev and ReTACRED. 

    \begin{table*}[]
        \centering
        \resizebox{0.8\textwidth}{!}{%
        \begin{tabular}{|c|ccc|ccc|ccc|}
        \hline
        Dataset & \multicolumn{3}{c|}{PALSTM} & \multicolumn{3}{c|}{CGCN} & \multicolumn{3}{c|}{SpanBERT-base-cased} \\ \hline
         & \multicolumn{1}{c|}{Precision} & \multicolumn{1}{c|}{Recall} & F1-Score & \multicolumn{1}{c|}{Precision} & \multicolumn{1}{c|}{Recall} & F1-Score & \multicolumn{1}{c|}{Precision} & \multicolumn{1}{c|}{Recall} & F1-Score \\ \hline
        TACRED & \multicolumn{1}{c|}{67.8} & \multicolumn{1}{c|}{64.6} & 66.2 & \multicolumn{1}{c|}{71.2} & \multicolumn{1}{c|}{62.6} & 66.7 & \multicolumn{1}{c|}{66.4} & \multicolumn{1}{c|}{66.1} & 66.3 \\ \hline
        TACRev & \multicolumn{1}{c|}{73.7} & \multicolumn{1}{c|}{74.8} & 74.3 & \multicolumn{1}{c|}{77.5} & \multicolumn{1}{c|}{72.6} & 75 & \multicolumn{1}{c|}{71.3} & \multicolumn{1}{c|}{75.6} & 73.4 \\ \hline
        ReTACRED & \multicolumn{1}{c|}{78.8} & \multicolumn{1}{c|}{80.1} & 79.4 & \multicolumn{1}{c|}{80.5} & \multicolumn{1}{c|}{80} & 80.2 & \multicolumn{1}{c|}{82.5} & \multicolumn{1}{c|}{82.3} & 82.4 \\ \hline
        TACRED-EN & \multicolumn{1}{c|}{73.4} & \multicolumn{1}{c|}{66.8} & 69.9 & \multicolumn{1}{c|}{74.4} & \multicolumn{1}{c|}{67} & 70.5 & \multicolumn{1}{c|}{77.2} & \multicolumn{1}{c|}{65.5} & 70.9 \\ \hline
        TACRED-ENP & \multicolumn{1}{c|}{74.9} & \multicolumn{1}{c|}{68.3} & 71.5 & \multicolumn{1}{c|}{75.2} & \multicolumn{1}{c|}{70.7} & 72.9 & \multicolumn{1}{c|}{80.5} & \multicolumn{1}{c|}{72.5} & 76.3 \\ \hline
        TACRED-RN & \multicolumn{1}{c|}{74.1} & \multicolumn{1}{c|}{67} & 70.3 & \multicolumn{1}{c|}{76} & \multicolumn{1}{c|}{67.1} & 71.3 & \multicolumn{1}{c|}{76.8} & \multicolumn{1}{c|}{64.5} & 70.1 \\ \hline
        TACRED-RNP & \multicolumn{1}{c|}{75.6} & \multicolumn{1}{c|}{68.3} & 71.8 & \multicolumn{1}{c|}{77.1} & \multicolumn{1}{c|}{68} & 72.3 & \multicolumn{1}{c|}{76.5} & \multicolumn{1}{c|}{68.8} & 72.5 \\ \hline
        \end{tabular}%
        }
        \caption{Comparison of baseline models' performance on different variants of TACRED.}
        \label{tab:perf-comp}
    \end{table*}

\textbf{Qualitative Comparison of IS and ES:} Our analysis in section \ref{sec:tacred} indicates that negative examples are the main source of noise in the dataset. So, we assumed that the majority of the positive instances are clean for the \textit{IS}, whereas no such assumptions were made for the \textit{ES}. \textit{IS} relies only on the model’s prediction, whereas \textit{ES} involves external help in obtaining the clean subset. As a result, \textit{IS} guarantees cost-efficiency whereas \textit{ES} ensures higher accuracy. Furthermore, using \textit{IS}, the same algorithm can be used for finding instances for elimination and reannotation. Whereas for \textit{ES}, we have separate algorithms for finding instances for elimination and reannotation.
    
\subsection{Analysis of Noisy Sample}
\label{sec:noisy_sam_analysis}
\noindent \textbf{Uncertainty Analysis: } To account for the uncertainty of deep learning models, we performed an experiment by replacing a \textit{linear} classifier layer with a \textit{Bayesian} classifier layer. The results from using the two baseline models are shown in Table~\ref{tab:bayesian}. As it can be seen, the two models gave different results. The performance of the PALSTM model deteriorated, while the CGCN model improved its performance in terms of F1-score by almost 1\%. One plausible explanation could be that the inconsistency in the results of the two models is due to \textit{aleatoric} uncertainty, i.e., uncertainty in the data \cite{uncertainty-review}. Data uncertainty is considered irreducible, as it corresponds to the inherent property of the dataset \cite{uncertainty-review}. Of course, this warrants further detailed analysis, which is beyond the scope of this paper.

    \begin{table*}[]
        \centering
        \resizebox{0.8\textwidth}{!}{%
        \begin{tabular}{c|c|c|c|c|c|c|c}
        Model & Classifier & Precision & Recall & F1-Score & Positive Accuracy & Negative Accuracy & Accuracy \\ \hline
        \multirow{2}{*}{PALSTM} & Linear & 66.5 & 65.7 & 66.1 & 65.7 & 92.2 & 86.5 \\  
         & \begin{tabular}[c]{@{}c@{}}Bayesian\\ Linear\end{tabular} & 68.3 & 62.9 & 65.5 & 62.9 & 93.2 & 86.7 \\ \hline
        \multirow{2}{*}{CGCN} & Linear & 71.6 & 60.8 & 65.8 & 60.8 & 94.5 & 87.3 \\  
         & \begin{tabular}[c]{@{}c@{}}Bayesian\\ Linear\end{tabular} & 70.6 & 63.5 & 66.8 & 63.5 & 93.8 & 87.3 \\ \hline
        \end{tabular}%
        }
        \caption{Baseline models' performance comparison. For each model, ``Linear" and ``Bayesian Linear" indicate the final layers as Linear and Bayesian Classifier respectively.}
        \label{tab:bayesian}
    \end{table*}

Furthermore, we also performed a simple analysis of the probability score assigned to the predicted relation label by the Bayesian classifier. The reported result is for the CGCN-Bayesian model (CGCN with Bayesian Linear final layer). However, similar results were observed for the PALSTM-Bayesian model (PALSTM with Bayesian Linear final layer). The probability score across all test instances ranges from 0.2 to 0.99. We consider examples with a probability score for the predicted class of less than 0.8 to be the ones with higher uncertainty. Such examples can be considered "noisy instances". We compared those instances with noisy instances selected either for elimination or for reannotation by the two proposed strategies. We found more than 40\% overlap of the noisy instances selected by either of the strategies with the uncertain examples detected by the CGCN-Bayesian model. This analysis indicates that future work can consider combining uncertainty measures and heuristics to obtain noisy instances of data.

\noindent \textbf{Manual Analysis: }
To evaluate the quality of examples shortlisted as noisy for elimination or reannotation following the extrinsic strategy, we randomly sampled a hundred examples that are common in identified potentially noisy examples by the PALSTM and CGCN models. We analysed those sentences based on the following error categories: (i) \textit{Falsely labeled as no\_relation}, (ii) \textit{Incorrect Span:} Entities span is not properly marked/identified, and (iii) \textit{Wrong Entity Type:} Entity type for one of the argument entities is incorrect. 

Out of 100 sampled examples for reannotation, 68 examples labeled with no\_relation are incorrectly labeled and they should be labeled with one of the 41 relation labels. 17 examples are falsely identified as noisy. The remaining can be attributed to the other two error categories: incorrect span and wrong entity type.

Out of 100 sampled for elimination, 62 examples labeled with no\_relation are incorrectly labeled. 15 examples are falsely identified as noisy and the remaining can be attributed to error categories such as incorrect span and wrong entity types.

    Despite falsely identifying a small percentage as noisy, our approach is capable of providing substantial noisy instances (68 out of 100). Thus, it can be easily scaled for any large dataset. 
\subsection{Robustness of our Approach}
    All the results discussed in the previous section are based on a single run of a model. Thus, to verify whether those results are not biased by the model parameters, we performed a small experiment. We trained PALSTM another four times with distinct initialization. For each distinct run, we generated corresponding sets of noisy instances for elimination and reannotation. In the comparison of these sets, we observed, that \textit{(i)} the number of instances generated for elimination and reannotation are in the range of 453 to 481 and 373 to 409 respectively. \textit{(ii)} more than 50\% of instances are common across all the sets for both reannotation and elimination, and \textit{(ii)} more than 70\% instances are identified as noisy by more than 3 different models.

The above discussion indicates that the proposed strategies are robust to multiple runs of a single model. Further, modifying the strategies to consider potential noisy instances based on multiple runs is likely to improve the model performance.

\subsection{Impact on different Relation Labels}
    The statistical impact of eliminating and reannotating noisy instances using clean data on the relation label set is presented in Table~\ref{tab:impact}. All the reported numbers are the intersection of two baseline models CGCN \cite{cgcn} and PALSTM \cite{tacred}. 11 relations have shown performance improvement and for 4 relation labels performance did not change across all the strategies. There is no common relation with declining performance across all 4 strategies.
    
    \subsubsection{Impact of eliminating noisy instances:} In our dataset analysis we have shown that negative instances are the main source of noise in the dataset and eliminating them has shown performance improvement across both our strategies (Tables~\ref{tab:is-elim-perf}, \ref{tab:es-elim-perf}). On evaluating the impact of removing negative noisy instances following our best strategy (ES), i.e. using a clean set, 20 relation labels show improved performance. Moreover, by eliminating noisy positive instances as well, the number of relation labels rises to 23. However, 17 relation labels are common to both the elimination strategies.
    
    \subsubsection{Impact of reannotating noisy instances:} Although reannotating noisy instances has not shown performance improvement across both our strategies (Tables~\ref{tab:is-reann-perf}, \ref{tab:es-reann-perf}), using clean dataset, the performance improvement is significant. On evaluating the impact of reannotating negative noisy instances following our best strategy, i.e. using a clean set, 18 relation labels show improved performance. Moreover, by eliminating noisy positive instances as well, the number of relation labels rises to 19. However, 15 relation labels are common to both the reannotation strategies.
    
    Out of 24 relation labels considered in a clean set other than no\_relation, on reannotating only negative instances, performance for 14 relations improved for both the models, and for 7 relations, it improved for at least one model. While on reannotating both positive and negative instances, for 17 relations performance of both the models improved, and for 4 relations performance improved for at least one model. The above observations establish that including unambiguous instances for relations not covered in the clean set can further improve the model's performance. 
    
    \begin{table}[]
        \centering
        \resizebox{0.5\textwidth}{!}{%
        \begin{tabular}{|c|c|c|c|}
        \hline
        \textbf{Strategy} & \textbf{\begin{tabular}[c]{@{}c@{}}\#Rel Performance \\ Improved\end{tabular}} & \textbf{\begin{tabular}[c]{@{}c@{}}\#Rel Performance \\ Declined\end{tabular}} & \textbf{\begin{tabular}[c]{@{}c@{}}\#Rel Performance \\ Remained Same\end{tabular}} \\ \hline
        Eliminating Negative & 20 & 1 & 7 \\ \hline
        Reannotating Negative & 18 & 4 & 5 \\ \hline
        \begin{tabular}[c]{@{}c@{}}Eliminating Negative \\ \& Positive\end{tabular} & 23 & 3 & 6 \\ \hline
        \begin{tabular}[c]{@{}c@{}}Reannotating Negative\\ \& Positive\end{tabular} & 19 & 5 & 6 \\ \hline
        \end{tabular}%
        }
        \caption{Impact of different strategies on performance of different relation labels. Each number represents the intersection of PALSTM and CGCN models.}
        \label{tab:impact}
    \end{table}

\subsection{Limitations}
    In the extrinsic strategy, we have used the ReTACRED test set as clean data to avoid any form of human intervention. However, to adopt our proposed method, one needs to employ expert annotators to obtain a high-quality clean subset of data. The clean subset can also be obtained by considering high-confidence examples from multiple baseline models. Nevertheless, the quality of the data can be questionable, thus it is left for investigation in future work.

    Further, the proposed strategies can over-clean the dataset as we have earlier shown that 17 out of 100 examples are falsely identified as noisy. However, we can control this to some extent by taking only the most confident noisy instances across multiple runs of a model. The use of quantitative uncertainty models can be an alternative way to control this as the uncertainty analysis in the section \ref{sec:noisy_sam_analysis} indicates.
    

\section{Conclusion \& Future Work}
\label{sec:conclusion}
This paper presented a model-based characterization of the noise present in the TACRED dataset and two strategies to handle potentially noisy instances. To the best of our knowledge, this is the first work that uses models' prediction and performance to characterize the noisy nature of the dataset. Moreover, this work can be easily automated and generalized for any other classification task. Analyses of the models' prediction results indicate that the incorrect labeling of instances as no\_relation class or negative relation contributes significantly to the noise in the data. Hence, this work proposes two different strategies for identifying potentially noisy negative relation instances for elimination and reannotation. The first strategy, the intrinsic strategy, is based on finding the nearest neighbor to the model's false negative prediction. Whereas, the second strategy, the extrinsic strategy, requires a subset of clean TACRED instances. Models trained on a dataset with elimination based on the intrinsic strategy show improvement when models are evaluated on the cleaner version of the test set. The performance of the models significantly improved with the extrinsic strategy for both the eliminated and reannotated datasets. Furthermore, identifying noisy instances among positive relation classes using the extrinsic strategy shows more improvement in models' performance.

Even though the proposed methodologies led to a performance improvement, they were able to identify only a small subset of noisy instances. Moreover, identifying all noisy instances without manual intervention is practically very challenging. Therefore, this work can further be extended by exploring approaches such as curriculum learning \cite{curriculumrc}, multi-network learning \cite{jo-src}, and label refurbishment \cite{chen2021beyond}. Another interesting future direction could be to explore the idea of using uncertainty quantification methods \cite{uncertainty-review} to identify potential noisy instances and utilize them for reannotation.


\bibliographystyle{ACM-Reference-Format}
\bibliography{sample-base}

\appendix

\end{document}